\documentclass[sigconf]{acmart}
\usepackage{tabularx}
\usepackage{graphicx}
\usepackage{booktabs}
\usepackage{makecell}
\usepackage{adjustbox}
\usepackage{array}
\usepackage{multirow}

\newcommand{\fm}[1]{\textcolor{blue}{fm: #1}}

\newcommand{\smstext}[1]{\textcolor{cyan}{sms: #1}}

\AtBeginDocument{%
  }

\setcopyright{none}
\copyrightyear{2024}
\acmYear{2024}
\acmDOI{XXXXXXX.XXXXXXX}
\acmConference[FAccTRec '24]{Make sure to enter the correct
  conference title from your rights confirmation emai}{October 14--18,
  2024}{Bari, Italy}
\acmISBN{978-1-4503-XXXX-X/18/06}

\begin{document}

\title{Towards Fairer Health Recommendations: finding informative unbiased samples via Word Sense Disambiguation}

\author[1]{Gavin Butts}
\authornote{Equal contribution.}
\affiliation{
  \institution{Loyola Marymount University}
  \city{}
  \country{}
}
\email{gbutts@lion.lmu.edu}

\author[2]{Pegah Emdad}
\authornotemark[1]
\affiliation{
  \institution{Worcester Polytechnic Institute}
  \city{}
  \country{}
}
\email{pemdad@wpi.edu}

\author[3]{Jethro Lee}
\authornotemark[1]
\affiliation{
  \institution{Northeastern University}
  \city{}
  \country{}
}
\email{lee.jet@northeastern.edu}

\author[4]{Shannon Song}
\affiliation{
  \institution{Worcester Polytechnic Institute}
  \city{}
  \country{}
}
\email{smsong@wpi.edu}


\author[6]{Chiman Salavati}
\affiliation{
  \institution{University of Connecticut}
  \city{}
  \country{}
}
\email{chiman.salavati@uconn.edu}

\author[4]{Willmar Sosa Diaz}
\affiliation{
  \institution{University of Connecticut}
  \city{}
  \country{}
}
\email{willmar.sosa\_diaz@uconn.edu}

\author[8]{Shiri Dori-Hacohen}
\affiliation{
  \institution{University of Connecticut}
  \city{}
  \country{}
}
\email{shiridh@uconn.edu}

\author[5]{Fabricio Murai}
\affiliation{
  \institution{Worcester Polytechnic Institute}
  \city{}
  \country{}
}
\email{fmurai@wpi.edu}




\renewcommand{\shortauthors}{Butts et al.}

\begin{abstract} 

    There have been growing concerns around high-stake applications that rely on models trained with biased data, which consequently produce biased predictions, often harming the most vulnerable. 
    In particular, biased medical data could cause health-related applications and recommender systems to create outputs that jeopardize patient care and widen disparities in health outcomes. A recent framework titled \emph{Fairness via AI} posits that, instead of attempting to correct model biases, researchers must focus on their root causes by using AI to debias data. Inspired by this framework, we tackle bias detection in medical curricula using NLP models, including LLMs, and evaluate them on a gold standard dataset containing 4,105 excerpts annotated by medical experts for bias from a large corpus. We build on previous work by coauthors which augments the set of negative samples with non-annotated text containing social identifier terms. However, some of these terms, especially those related to race and ethnicity, can carry different meanings (e.g., ``\emph{white} matter of spinal cord''). To address this issue, we propose the use of Word Sense Disambiguation models to refine dataset quality by removing irrelevant sentences. We then evaluate fine-tuned variations of BERT models as well as GPT models with zero- and few-shot prompting. We found LLMs, considered SOTA on many NLP tasks, unsuitable for bias detection, while fine-tuned BERT models generally perform well across all evaluated metrics.
    

\end{abstract}




\keywords{medical text data, bias detection, LLMs, word sense disambiguation}



\maketitle

\section{Introduction} 
\label{introduction}


 For decades, medicine has been marred by implicit and explicit biases that continue to negatively impact patient outcomes by perpetuating stereotypes and contributing to health disparities among social groups that face systemic oppression \cite{corsino2021impact, dehon2017systematic}. Despite efforts to remediate and address these biases from their source, many medical schools still incorporate biased medical teachings during the preclinical years \cite{halman2017using,tsai2016race}. Many educators continue to misuse race as a substitute for genetics or ancestry, or they use gender and sex terms incorrectly reinforcing the notion that sex and gender are binary or fixed rather than fluid, which can potentially alienate gender-nonconforming students and patients \cite{ali2011use, hunt2013genes,karani2017commentary}. The current focus in AI research is primarily on identifying and exposing bias within AI systems, often without addressing the root causes of bias inherent in the data these systems are built upon. As long as structural inequalities exist in the real world, AI systems will perpetuate these biases \cite{dori2021fairness}. 
  By harnessing machine learning to analyze and detect these biases, we can advance equity in medical training and the fairness of AI models, leading to a more accurate and effective healthcare system.

Recently, Salavati et al.~\cite{salavati2024reducing} introduced the BRICC (Bias Reduction in Curricular Content) dataset and proposed a systematic and scalable AI-based method for identifying potential bias in medical curricula. Given the steep cost of false negatives (i.e., classifying a biased sentence as unbiased), they emphasize that recall must be prioritized over precision. Moreover, due to the inherent difficulty of the task, one of the best approaches in this High-Recall Information Retrieval setting is the Technology Assisted Review (TAR)~\cite{cormack2016engineering,kusanormalised}, whereby a set of experts reviews the samples flagged by a model, as envisioned by Salavati et al. The paper also used a curated list of social identifiers to find additional, negative (i.e., non-biased) samples in unlabeled data. However, this latter approach suffered from social identifier terms that had ambiguous meanings, leading to lower quality of the training data, and negative samples that were too ``easy'' to classify as non-biased. For example, one social identifier used to filter for race-related data was ``white''. In Table~\ref{tab:race-wsd}, simply searching for the keyword ``white'' will include both race-related and non-race-related text excerpts. 

We believe that using these ambiguous meaning terms as-is leads to overestimating the true discernment power of the bias classifier by making the problem too easy. It may be that the classifier is actually differentiating race-related from non-race-related terms, rather than biased from non-biased sentences. 

For this reason, we propose a new framework to augment the sampling process for negative examples, using Word Sense Disambiguation (WSD) methods for data enhancement. We hypothesize that this would improve the distinction between biased and non-biased sentences in the bias classification process. 






\begin{table} 
\centering
\small
\caption{The term ``white'' in a racial vs. non-racial context}
\begin{tabularx}{0.95\columnwidth} { 
  >{\arraybackslash}X 
  | >{\arraybackslash}X }
 \toprule
     \multicolumn{1}{c|}{\textbf{Race-Related}} & \multicolumn{1}{c}{\textbf{Not Race-Related}}\\
     \midrule
      ``5 Year Relative Survival: overall 84\% for \textbf{white} women, 62\% for black women, 95\% for local disease, 69\% regional disease (spread to lymph node), 17\% for distant disease.'' & ``\textbf{White} matter within the spinal cord contains the axons of neurons that are ascending and descending to transmit signals to and from the brain, respectively.'' \\
     \bottomrule
    \end{tabularx}
    \label{tab:race-wsd}
\end{table}




Our main contributions are as follows:
\begin{enumerate}

    \item We enhance a framework for detecting bias in medical curriculum content, with a focus on improving data quality.
    
    \item We leverage Word Sense Disambiguation (WSD) models in training bias detection classifiers by filtering out irrelevant samples from the data. Moreover, we use ChatGPT-4o to augment a set of manually labeled examples with synthetic samples to fine-tune and/or evaluate WSD models.
    
    \item We fine-tune and evaluate various Transformer-based models including DistilBERT, RoBERTa, and BioBERT for the bias detection task. In addition, we use Large Language Models (LLMs), such as TinyLlama, for bias classification, evaluating zero-shot vs.\ few-shot prompting with GPT, to improve the performance of bias detection.
    
    \item We present a comprehensive evaluation of the various models, highlighting the improvements achieved through the use of WSD and ChatGPT-generated sentences.
    
\end{enumerate}


\section{Related Work} \label{related work}

\subsubsection*{Health Recommender Systems (HRS)} 
Recommender systems have become integral to the healthcare industry, providing personalized medical recommendations that enhance patient understanding of their medical condition and improve health outcomes~\cite{tran2021recommender}. 
These systems assist healthcare professionals in predicting and treating diseases by analyzing patient data to recommend personalized diets, exercise regimens, medications, diagnoses, and other health services~\cite{sahoo2019deepreco,pincay2019health}. 
Despite numerous studies exploring various aspects of HRS, the literature for addressing various types of biases in such systems rooted in curricula contents is limited~\cite{salavati2024reducing}.
\subsubsection*{Debiasing medical corpora manually and via AI} 
Concerns over biased AI models and, particularly, recommender systems in healthcare applications have been gaining more attention due to their increased use in high-stake decisions~\cite{challen2019artificial}. In essence, their biases are rooted in implicit and explicit biases embedded in the data used for training them \cite{howcroft2019bias}
, which stem from various sources, including inherent biases in medical literature, the subjectivity of human annotators, and historical and systemic inequities present in healthcare systems~\cite{salavati2024reducing}. 
Numerous recent studies have aimed to quantify and address this issue both manually and through AI. Khan et al.~\cite{khan2023gender} manually explored the systemic bias held by medical professionals when writing recommendation letters. On the other hand, Raza et al.~\cite{raza2024nbias} and Salavati et al.~\cite{salavati2024reducing} aimed to detect bias in medical text using transformer-based language models. The former used a semi-autonomously labeled dataset covering diverse medical topics, whereas the latter employed a dataset manually labeled by medical experts focusing on biased information in medical curricular texts. Although both studies provide a comprehensive overview of bias detection, ensuring high-quality data remains an issue. While we also explore AI models for debiasing medical text data, we investigate better ways of augmenting the set of unbiased samples and consider a wider gamut of models, including LLMs.



\subsubsection*{Machine Learning for Bias Detection} 
%
%
Prior works have applied various BERT models for bias classification tasks. Tiderman et al.~\cite{reu2023} used DistilBERT, a transformer-based distilled BERT model, to classify biased information in social media content. Similarly, Raza et al.~\cite{raza2024nbias} achieved the best results for bias classification in medical text through fine-tuning BERT, a simple encoder-only transformer. 
Building on the existing literature for bias detection in medical contexts, we additionally apply Large Language Models (LLMs) for this task. Specifically, we use TinyLlama \cite{zhang2024tinyllama}, a computationally efficient variant of Llama 2, for bias classification. In addition, we consider additional strategies for constructing the set of negative samples-- such as through the use of WSD for data refinement.

\subsubsection*{Use of LLMs for NLP tasks and prompt engineering} 
In NLP tasks, prompting LLMs have been shown to perform on par with encoder-only architectures, like BERT, without the need for fine-tuning~\cite{colavito2024leveraging}. It has been shown that prompting techniques, such as zero-shot, few-shot, or chain of thought (CoT), serve a key role in the quality and correctness of a model's output~\cite{naveed2023comprehensive}. 
%
These techniques have been used in many tasks, such as sentiment analysis~\cite{bu2023efficient}, text classification~\cite{clavie2023large}, as well as for healthcare applications, such as question-answering, and as a clinical recommender system~\cite{wang2023prompt,patil2024prompt}. Despite these initiatives, we are the first to evaluate zero- and few-shot prompting for detecting bias in medical curricular content.

\section{Dataset}
\label{sec:dataset}
Our work builds on the BRICC dataset introduced by Salavati et al.~\cite{salavati2024reducing}, which consists of 509 PDF files and 12,647 pages of medical school instructional materials annotated by medical students and experts trained in identifying bias. Within the dataset, there are three tiers of coding. The first-level codes identify social identifiers within the excerpt. The second-level codes assess the presence or absence of bias in the excerpt, categorized into four distinct groups: `\textbf{biased}', `\textbf{potentially biased}', `\textbf{non-biased}', and `\textbf{review}'. Additionally, third-level codes establish a link between a medical condition and one or more categories of social identifiers (e.g., race), specifying the type of identity and whether it was portrayed in a biased or unbiased manner. Each excerpt is then assigned one or more codes formatted as ``TYPE-disease'', where TYPE represents one of 17 categories of social identifiers. Akin to the previous work, we focus on the most frequent types including sex, gender, race, ethnicity, age, and geography. Each category is associated with a list of keywords that can signify social identifiers.

\subsubsection*{Positive and negative samples.} \textbf{Positive samples} are defined as those excerpts that contain either a `biased', `potentially biased', or `review' and a selected ``TYPE-disease''. \textbf{Negative samples} are subdivided into various types, as detailed in the previous work~\cite{salavati2024reducing}. The negative types that we are most interested in are referred to as extracted negatives (XN). In this case, these are sentences from the corpus that, despite containing a category keyword, were deliberately excluded from the annotation process. Please refer to supplemental materials for a detailed explanation of negative types found in the BRICC dataset. Of the XN types, we filtered for those that contained at least one relevant keyword relating to our selected ``TYPE-disease''. The distribution of positives and negatives across the dataset is displayed in Table~\ref{tab:bricc}.


\begin{table}[tb]
    \centering
    \caption{BRICC Dataset Characteristics}
    \begin{tabular}{lr}
        \hline
        & \textbf{Counts} \\
        \hline
        Number of PDF Files & 509 \\
        Total Number of Pages & 12,647 \\
        Annotated Excerpts & 4,105 \\
        Labeled Positives & 1,116 \\
        Labeled Negatives (LN) & 2,989 \\
        Extracted Negatives (XN) & 4,391 \\
        \hline
    \end{tabular}
    \label{tab:bricc}
\end{table}

We focus on XN samples in our experiments because the authors previously reported higher recall, 0.925, but at the expense of precision, 0.504, by using this negative set~\cite{salavati2024reducing}. Despite this improved performance, we noticed that many of the keywords may lead to the inclusion of non-``TYPE-disease'' related content. An example of this occurrence may be seen in Table~\ref{tab:race-wsd}. Hence, retaining only negative samples that relate to the social demographics of interest is a key computational task, which we address using WSD. 

\subsubsection*{Labeling data for WSD} To gather samples suitable for training WSD models, we selected those XN excerpts that contained a keyword related to the selected social demographics categories: sex, gender, race, ethnicity, geography, and age. After obtaining a random sample of XN excerpts for each category, we had a human expert annotate whether the meaning of the keyword term was indeed related to that category or not. Based on the results of this annotation process, we decided to only focus on race keywords because they suffered the most from ambiguity. The other bias categories did not have a significant degree of ambiguity that required correction. These labeled excerpts were used to train our WSD models. 

\section{Preliminaries}


\subsection{LLM Prompting}
Reynolds et al.~\cite{reynolds2021prompt} suggest that zero-shot prompts could significantly
outperform few-shot prompts. Their analysis highlights the
need to consider the role of prompts in controlling and evaluating
the performance of language models. Their study stated that since
GPT-3 is often not learning from few-shot examples during the run
time, this model can effectively be prompted without examples \cite{reynolds2021prompt}.
Additionally, Kojima et al.~\cite{kojima2022large} demonstrate that chain of thought (CoT) prompting, a recent technique for eliciting complex multi-step reasoning through step-by-step answer examples, achieved state-of-the-art performances in arithmetics and symbolic reasoning tasks. They proposed Zero-shot-CoT, a zero-shot template-based prompting using chain of thought reasoning, and highlighted its high performance.



\subsection{Word Sense Disambiguation Task Definition}

Generally speaking, word sense disambiguation (WSD) is the task of identifying the correct sense of a polysemous $w \in \mathcal{W}$ (a word with multiple meanings) in a given context $x$. Formally, given a set of words $\mathcal{W}$, a finite set of possible senses $\mathcal{S}_w = \{S_w^{(1)}, \ldots, S_w^{(k)}\}$ for each $w \in \mathcal{W}$ and a context (ordered sequence of words) $x = (x_1, \ldots, x_{i-1}, w, x_{i+1},\ldots\,x_n) \in \mathcal{X}$, find a function $f: \mathcal{W} \times \mathcal{X} \rightarrow \mathcal{S}$, such that $f(w,x)$ is the correct sense of $w$ in context $x$.

For this paper, we are interested in determining if a term $w$, listed as a possible social identifier for category $t$, is related to $t$ in an excerpt $x$. To do so, we need to learn a function $\textsc{IsRelated}(w,x,t) \in \{\textsc{true}, \textsc{false}\}$.
\sloppy For instance, the set of social identifiers for 
race is $\mathcal{S}_{\textrm{race}} = \{\textrm{`white'}, \textrm{`black'}, \ldots\}$. Ideally, in one of the examples seen earlier, we want $\textsc{IsRelated}(\textrm{`white'},\textit{`white matter within...'},\textrm{`race'}) = \textsc{false}$.




\subsection{Bias Detection Task Definition}
In the context of medical education, we consider bias detection as a High Recall Information Retrieval task. It is the first step in a TAR system. This task consists of classifying a text excerpt $x$ as unbiased ($\hat y = 0$) or potentially biased  ($\hat y = 1$). In the latter case, the sample would be subsequently reviewed by a medical expert.


Bias may be related to one or more categories of social identifiers, including race, ethnicity, sex, gender, age, and geography. For instance, ``\textit{They promote hair growth in the groin, axilla, chest and face, yet they also promote hair loss in the scalp in men who are genetically susceptible to androgenetic alopecia.}'' is labeled by medical excerpts as `\textbf{biased}' with respect to \textit{gender} (designated by the social identifier \textit{men}). As explained, in the comment from one of the annotators: ``\textit{Use sex terms when speaking of populations, should be male instead of men. Also, include citation to support this assertion.}''

Formally, Salavati et al.~\cite{salavati2024reducing} define type-specific bias as a binary label $\textsc{bias}(x, t) \in \{\textsc{true}, \textsc{false}\}$ indicating whether excerpt $x$ is biased with respect to a social identifier category $t$. In the present work, we consider only the \emph{general} definition of bias, regardless of which category $t$ in a set $\mathcal{T}$ it belongs to:
$
\textsc{bias}(\textit{x}, \mathcal{T} ) = \textsc{true} \iff \exists t \in \mathcal{T} \text{ s.t. } \textsc{bias}(\textit{x}, \textit{t}) = \textsc{true}.
$

\section{Methodology}

\begin{figure}[tb!]
\includegraphics[width=0.5\textwidth]{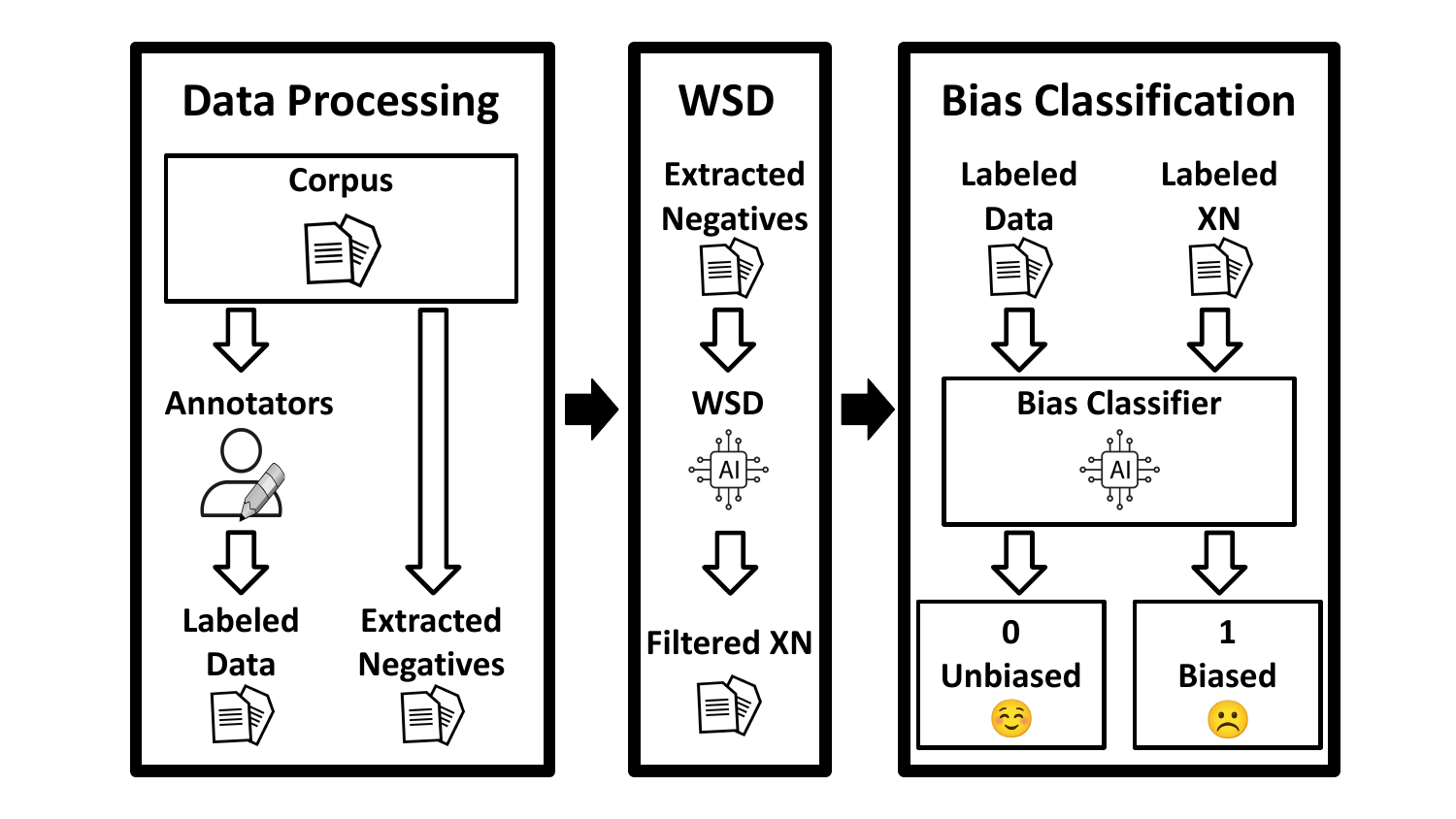}
    \caption{Workflow stages. (Left) Data processing: annotated excerpts are labeled as `biased' (positive) or `non-biased' (negative); XN: additional sentences extracted as negative examples. (Center) Word Sense Disambiguation (WSD) used for selecting from XN relevant negatives (XN$^*$). (Right) Training and evaluation of bias classifiers.}
    \label{fig:overview}
\end{figure}

In this section, we provide an overview of the proposed framework. Figure~\ref{fig:overview} (left) shows the data processing steps performed by Salavati et al.~\cite{salavati2024reducing}, which we leverage in the present work. As explained, in addition to labeled data, BRICC includes negative samples extracted from the non-annotated data (denoted as XN). Figure~\ref{fig:overview} (center) illustrates the application of WSD to filter out irrelevant samples, which results in the filtered XN set (XN$^*$). This process is described in detail in Section~\ref{sec:WSD}. Last, Figure~\ref{fig:overview} (right) depicts the augmentation of labeled data with XN$^*$ for training different bias classifiers, whose performance we evaluate. Details are discussed in Section~\ref{sec:bias methodology}.




\subsection{Word Sense Disambiguation Experiments} \label{sec:WSD}

We evaluate several models for WSD: a simple baseline, two fine-tuned variants of BERT, and two GPT models. For fine-tuning and evaluation, we combined our manual annotations with sentences generated by ChatGPT-4o, yielding 352 labeled excerpts. 

\begin{figure}[tb!]
    \centering
    \includegraphics[width=0.5\textwidth]{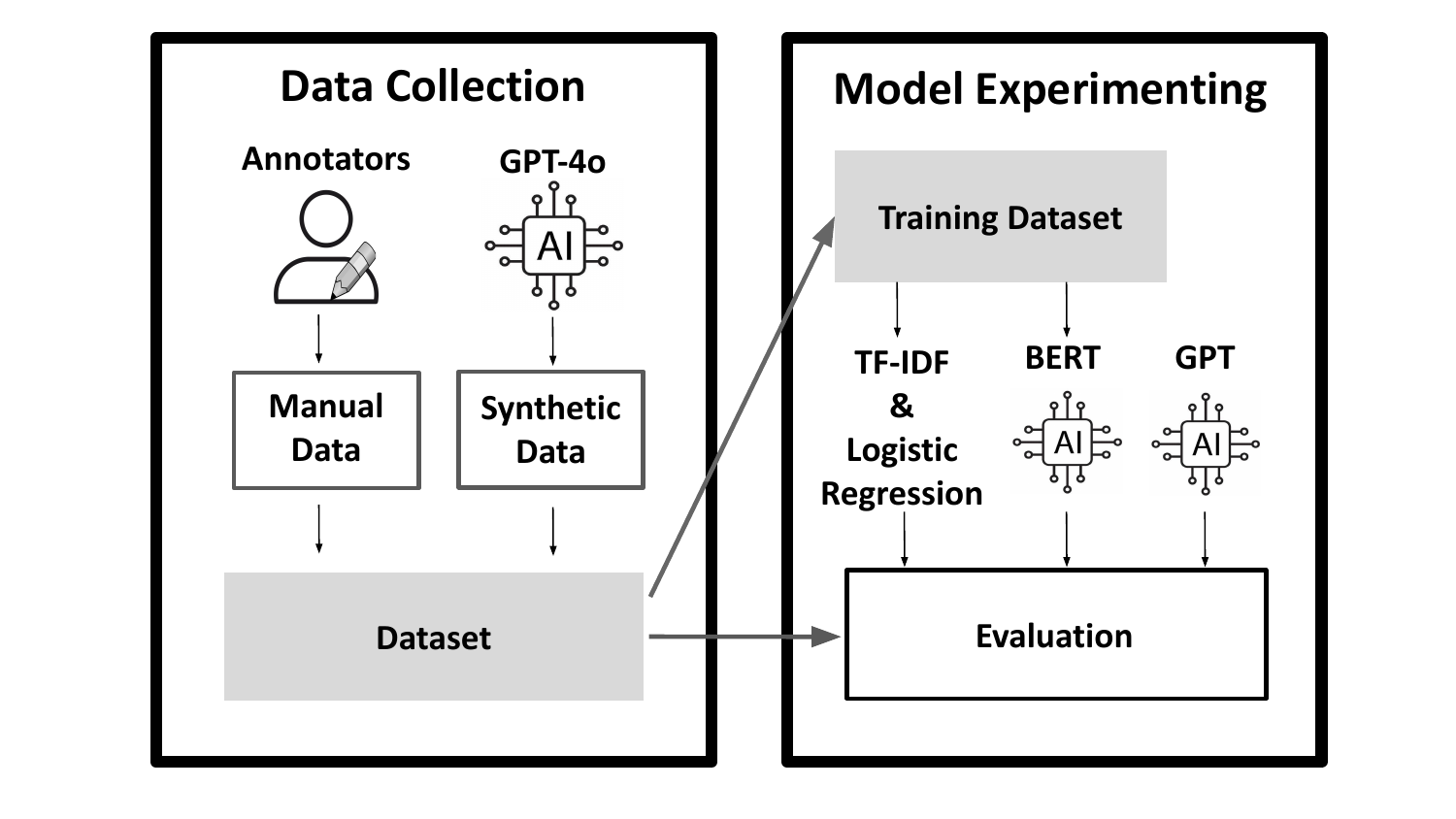}

    \caption{WSD training and evaluation. Excerpts manually labeled as race-related or not plus GPT-generated sentences are used to train and evaluate the WSD models.}

    \label{fig:wsd-overview}
\end{figure}




\subsubsection*{Experimental Setup.} 
We divide each of the two datasets (manually annotated excerpts and synthetic samples) independently with a 70-15-15 stratified split into training, validation, and test sets.

We investigated two ways of building the training set:
\begin{itemize}
    \item Only manually-annotated excerpts;
    \item Both manually-annotated excerpts and synthetic samples.
\end{itemize}


\subsubsection*{WSD Models} 
We evaluate three models shown in the recent literature to perform well on WSD: ALBERT, GlossBERT, and GPT models. These models are compared to our baseline, a logistic regression with TF-IDF.

\subsubsection*{Fine-tuning and Prompting.}
We fine-tune all layers of the pre-trained ALBERT and GlossBERT models with a learning rate of $2\times 10^{-5}$ and weight decay of $0.01$ over $10$ epochs, keeping the model that yielded the smallest validation loss.

Given the substantial empirical and theoretical evidence supporting the benefits of chain of thought (CoT) prompting in various LLM tasks~\cite{cot2022neurips,cot2023neurips}, we incorporate CoT into our zero-shot prompts. We opted to follow the prompt template presented in Kojima et al.~\cite{kojima2022large}, which was shown to produce the highest accuracy (i.e., ``Let’s think step by step''). To prompt the model, we first specified the model’s role: ``\textit{You are a helpful assistant that determines
if the sentence is race or ethnicity related}''. Then, we defined the task as: ``\textit{Given the sentence `{text}', think step by step: Is this sentence race or ethnicity related? Only output 1 or 0. If this sentence contains any terms relating to race or ethnicity, state 1. Otherwise, state 0.}'' 




\subsubsection*{Metrics.} We evaluate the models' performance on the test set with respect to accuracy, precision, recall and  F1 score. 

\subsection{Bias Classification Experiments}
\label{sec:bias methodology}

Using the BRICC dataset, we fine-tune binary classification neural language models and prompt pre-trained LLMs for bias classification. For fine-tuning, we consider encoder-only and decoder-only models (encoder-decoder models are often reserved for multi-modal tasks and causal language inference~\cite{asadi2020encoder}).
The fine-tuned models include RoBERTa, DistilBERT, BioBERT, and TinyLlama. Using prompt engineering, we additionally prompt GPT-4o mini.

\subsubsection*{Different sets of negatives.} The datasets we will consider contain all positive samples, plus one of the following:
\begin{itemize}
    \item Labeled negatives (LN),
    \item Labeled negatives plus extracted negatives filtered by keywords (LN + XN), and
    \item Labeled negatives plus extracted negatives filtered using word sense disambiguation (LN + XN$^*$).
\end{itemize}
As described in Section~\ref{sec:dataset}, the set of extracted negatives (XN) is constructed by filtering data based on keywords that relate to either gender, sex, race, ethnicity, age, and/or geography. Then, we apply the best-performing WSD model to ensure these samples truly relate to the social demographics of interest, resulting in XN$^*$. To assess performance variability as a function of the data splits, we split the dataset in K-folds for cross-validation and calculate average performance and confidence intervals.



\subsubsection*{Fine-tuning.}  To construct the models, we add a classification head to each language model and fully fine-tune each model along with the classification head. The models we utilize are: RoBERT, DistilBERT, and BioBERT, all encoder-only, and TinyLlama, a 1.1B decoder-only model derived from Meta's Llama~2.

For each dataset we outlined, we do initial fine-tuning on RoBERTa, DistlBERT, BioBERT, and TinyLlama with a batch size of $8$ and a learning rate of $2\times10^{-5}$. We use the validation set to tune the hyperparameters with grid search, leading to the final model.

\subsubsection*{Prompting.} We evaluate the performance of GPT-4o mini by using zero- and few-shot prompting. To prompt this model, we first establish the model's role: ``\emph{You are a helpful assistant that determines if text is biased}''. Then, we establish the task. We find that the best task description is:
%
    ``\emph{Given text, determine if the text contains bias or no bias. 
         The bias may target gender, sex, race, ethnicity, age, and/or geography, include exclusive language, or make unsupported claims.
         The text may also contain no bias at all.
         If the text has bias state 1, if the text does not have bias state 0.}''

Both zero- and few-shot prompting used the prompt above. For few-shot prompting, we tested various sets of examples from the dataset. Table~\ref{table:few-shot prompting} shows the subset that performed best.

\subsubsection*{Metrics.} For the bias detection task, we also evaluate the models' performance on the test set with respect to precision, recall and F1 score. In addition, we consider the F2 score and area under the ROC curve (AUC). F2 is similar to F1 but prioritizes recall over precision. Due to the class imbalance, AUC is more relevant than accuracy because it accounts for all possible threshold choices. 

\begin{table}[tb!]
\centering
\small
\caption{Few-shot example inputs, outputs, and reasoning used for prompting GPT-4o mini for bias classification}
\begin{tabular}{p{5.1cm} p{2.3cm}}
\toprule
\textbf{Input} $x$ & \textbf{Label, Comment} \\ \midrule
52 year old, married female with one daughter, employed as a school administrator with no prior psych history reports 2 month h/o [history of] sadness, subjective anxiety and intermittent trouble falling asleep. & \textbf{Label: 1}, Use gender terms like woman for case studies  \\ \midrule
Once patient is on another treatment for her disorder, she no longer needs this medication. & \textbf{Label: 0}, n/a (from XN set) \\ \midrule
Recent meta-analysis suggested no difference in prevalence among countries, rate is 1-2\% with increase during late adolescence. & \textbf{Label: 1}, Term `Late Adolescence' is an unclear time period \\ \bottomrule
\end{tabular}

\label{table:few-shot prompting}

\end{table}

\section{Results}

\subsection{Evaluation of WSD models}
\label{sec:wsd_results}

\begin{table}[tb!]
    \centering
    \caption{Performance metrics for WSD on manually-annotated+GPT excerpts. Best result for each metric shown in \textbf{bold}. GlossBERT and GPT-4o are tied as the best models.
    }
    \begin{adjustbox}{max width=\textwidth}
    \begin{tabular}{@{}lccccc@{}}
        \toprule
        \textbf{Metric} & \makecell{\textbf{TF-IDF+}  \\ \small{Logistic Reg.}} & \textbf{ALBERT} & \makecell{\textbf{Gloss}\\ \textbf{BERT}} & \makecell{\textbf{GPT-3.5} \\ \textbf{Turbo}} & \makecell{\textbf{GPT-4o} \\ \textbf{mini}} \\
        \midrule
        Accuracy & 0.839 & 0.926 & \textbf{0.944} & 0.925 & \textbf{0.944} \\
        Precision & 0.816 & 0.935 & \textbf{0.936} & 0.916 & \textbf{0.936}\\ 
        Recall & 0.839 & 0.977 & \textbf{1.000} & \textbf{1.000} & \textbf{1.000}\\
        F1 Score & 0.817 & 0.956 & \textbf{0.967} & 0.956 & \textbf{0.967} \\
        \bottomrule
    \end{tabular}
    \end{adjustbox}
    \label{tab:wsd_metrics}

\end{table}

Table \ref{tab:wsd_metrics} presents the evaluation results for the WSD models examined. The baseline achieved worse results than the other models for every metric. We find that GlossBERT outperforms ALBERT and that GPT-4o mini improves upon GPT-3.5 Turbo. Furthermore, both GlossBERT and GPT-4o are tied as the best models, both exhibiting a very high F1 Score (0.967). Using the cost as a tie-breaker between the two, we opted to use GlossBERT for the WSD task performed on the extracted negatives. 

\begin{table}[tb!]
    \centering
    \small
    \caption{Examples of WSD test cases and GlossBERT predicted probabilities for $y=1$. Each excerpt has a term (bolded) listed among race/ethnicity keywords.}
    \begin{tabular}{ m{23.5em} c<{\centering}}
        \toprule
        \textbf{Input} $x$ (label $y$) & \textbf{Prediction} \\
        \midrule
        Melanoma: increasing in incidence in the \textbf{white} population (CDC). ($y=1$) &  0.9998 \\
        \midrule
        2015 \textbf{American} Heart Association guidelines suggest treating patients presenting with systolic  BP above 150-220 mmHg, but they do not offer a specific BP target. ($y=0$) &  \textcolor{red}{0.9998} \\
        \midrule
        Calcific plaques are chalky \textbf{white} and arise from cardiac (aortic and mitral) valves. ($y=0$) & 0.0001 \\
        \bottomrule
    \end{tabular}
    \label{tab:text_classification_results}
\end{table}

Table~\ref{tab:text_classification_results} illustrates examples from the test set, two of which were correctly identified and one that was not. While the first and third were correctly predicted with high confidence, the middle row was incorrectly classified with a ``high confidence prediction''. 
For the few instances that GlossBERT incurred false positives, a closer inspection has revealed that those excerpts may be lacking enough context for this specific task. For example, in the middle row, the use of `American' only indirectly relates to the ethnicity of the people that an organization serves.

We also investigate whether the synthetic samples generated by ChatGPT-4o were trivial, which would artificially inflate performance. When we evaluate the model results on only manually annotated excerpts, the performance of all models stays somewhat similar, except for GPT models, both of which achieve 100\% accuracy. Therefore, the synthetic examples are at least as hard as the manually annotated excerpts for BERT models, justifying their use in our evaluation.

In addition, we evaluate the models' performance when trained only on the manually-annotated data. In this case, there is a performance drop for the fine-tuned models (ALBERT declines from \textbf{0.926} to \textbf{0.852} and GlossBERT declines from \textbf{0.944} to \textbf{0.852} accuracy), indicating that the synthetic samples help the models to generalize better. Furthermore, GlossBERT remains tied as the best model, which supports our choice of using it for building the set of filtered extracted negatives in the bias detection task.

\subsection{Evaluation of Bias Detection Models}


\begin{table}[tb!]
    \centering
    \caption{Performance Metrics and 95\%-CIs for Fine-Tuned Models trained on LN+XN* data. RoBERTa yields the highest averages, but it is statistically tied with DistilBERT.}
    \begin{tabular}{@{}lcccc@{}}
        \toprule
        \textbf{Metric} & \textbf{RoBERTa} & \textbf{DistilBERT} & \textbf{BioBERT} \\
        \midrule
        Precision       & 0.613 $\pm$ 0.015 & 0.605 $\pm$ 0.013 & 0.581 $\pm$ 0.014 \\
        Recall          & 0.692 $\pm$ 0.024 & 0.649 $\pm$ 0.030 & 0.620 $\pm$ 0.019 \\
        F1 Score        & 0.650 $\pm$ 0.014 & 0.626 $\pm$ 0.018 & 0.599 $\pm$ 0.010 \\
        F2 Score        & 0.674 $\pm$ 0.019 & 0.639 $\pm$ 0.025 & 0.611 $\pm$ 0.014 \\
        AUC             & 0.927 $\pm$ 0.003 & 0.921 $\pm$ 0.006 & 0.904 $\pm$ 0.003 \\
        \bottomrule
    \end{tabular}

    \label{tab:BERT results}
\end{table}

\begin{table}[tb!]
    \centering
    \caption{Performance Metrics and 95\%-CIs for Prompting GPT-4o mini. Best results for each metric shown in bold. AUC was ommitted as it cannot be computed for binary outputs.}
    \begin{tabular}{@{}lccc@{}}
        \toprule
        \textbf{Metric} & \textbf{Zero-Shot} & \textbf{Few-Shot} \\
        \midrule
        Precision       & \textbf{0.367 $\pm$ 0.071} & 0.259 $\pm$ 0.019 \\
        Recall          & 0.260 $\pm$ 0.029 & \textbf{0.610 $\pm$ 0.026} \\
        F1 Score        & \textbf{0.303 $\pm$ 0.040} & \textbf{0.363 $\pm$ 0.023} \\
        F2 Score        & 0.274 $\pm$ 0.032 & \textbf{0.480 $\pm$ 0.025} \\
        \bottomrule
    \end{tabular}

    \label{tab:bias prompting results}
\end{table}

\begin{table*}[tb!]
\centering
\caption{Performance metrics and 95\%-CIs for RoBERTa, TinyLlama trained on dataset variants (LN+XN*, LN+XN, LN). Best results among each model variants (resp.\ across all models) and statistical ties shown are bolded (resp.\ underlined).}
\begin{tabular}{l|ccc|ccc}
\toprule
\multirow{2}{*}{\textbf{Metric}} & \multicolumn{3}{c|}{\textbf{RoBERTa}} & \multicolumn{3}{c}{\textbf{TinyLlama}}  \\
                                 & \textbf{LN+XN*} & \textbf{LN+XN} & \textbf{LN} & \textbf{LN+XN*} & \textbf{LN+XN} & \textbf{LN} \\
\midrule
Precision                        & \textbf{0.613 $\pm$ 0.015} & \textbf{0.640 $\pm$ 0.021} & 0.526 $\pm$ 0.029 & \underline{\textbf{0.675 $\pm$ 0.008}} & \underline{\textbf{0.693 $\pm$ 0.028}} & 0.536 $\pm$ 0.020 \\
Recall                           & \textbf{0.692 $\pm$ 0.024} & 0.667 $\pm$ 0.023 & \underline{\textbf{0.719 $\pm$ 0.026}} & \textbf{0.548 $\pm$ 0.030} & 0.519 $\pm$ 0.029 & \textbf{0.607 $\pm$ 0.035}  \\
F1 Score                         & \underline{\textbf{0.650 $\pm$ 0.013}} & \underline{\textbf{0.652 $\pm$ 0.017}} & 0.606 $\pm$ 0.017 & \textbf{0.604 $\pm$ 0.021} & \textbf{0.593 $\pm$ 0.017} & \textbf{0.568 $\pm$ 0.016}\\
F2 Score                         & \underline{\textbf{0.674 $\pm$ 0.019}} & \textbf{0.661 $\pm$ 0.016} & \textbf{0.669 $\pm$ 0.016} & \textbf{0.569 $\pm$ 0.027} & \textbf{0.546 $\pm$ 0.024} & \textbf{0.591 $\pm$ 0.025}  \\
AUC                              & \underline{\textbf{0.927 $\pm$ 0.003}} & \underline{\textbf{0.930 $\pm$ 0.009}} & 0.910 $\pm$ 0.008 & \textbf{0.907 $\pm$ 0.005} & \textbf{0.903 $\pm$ 0.005} & 0.871 $\pm$ 0.011  \\
\bottomrule
\end{tabular}
\label{tab:llm_wsd_results}
\end{table*}

\begin{table}[tb!]
    \centering
    \caption{Performance Metrics and 95\%-CIs for Fine-Tuned Models against Baseline ($^*$Salavati et al., 2024). Best results and statistical ties shown in bold. }
    
    
    \begin{tabular}{lccc}
        \toprule
        \textbf{Metric} & \textbf{RoBERTa} & \textbf{TinyLlama} & \textbf{Baseline$^{*}$} \\
        \midrule
        Precision       & 0.613 $\pm$ 0.015 & \textbf{0.675 $\pm$ 0.008} & 0.504 $\pm$ 0.054\\
        Recall          & 0.692 $\pm$ 0.024 & 0.548 $\pm$ 0.030 & \textbf{0.812 $\pm$ 0.069}\\
        F1 Score        & \textbf{0.650 $\pm$ 0.014} & 0.604 $\pm$ 0.021 & 0.615 $\pm$ 0.022\\
        F2 Score        & 0.674 $\pm$ 0.019 & 0.569 $\pm$ 0.027 & \textbf{0.717 $\pm$ 0.027}\\
        AUC             & \textbf{0.927 $\pm$ 0.003} & 0.907 $\pm$ 0.005 & \textbf{0.923 $\pm$ 0.004}\\
        \bottomrule
    \end{tabular}

    \label{tab:full-scope results}
\end{table}

Firstly, we compare the performance of the fine-tuned BERT variants on the bias detection task.
Table \ref{tab:BERT results} displays the models' performance with respect to precision, recall, F1 and F2 score, and AUC. While RoBERTa and DistilBERT are statistically tied, BioBERT clearly performs worst among all BERT models. We select the RoBERTa model for further comparison due to the model's higher mean evaluation metrics with lower standard deviations.

Secondly, we evaluate the performance of zero- and few-shot prompting with GPT-4o mini on bias detection. Table~\ref{tab:bias prompting results} shows the results obtained using the prompting techniques outlined in Section~\ref{sec:bias methodology}. Despite the substantial increase in recall seen with few-shot prompting, the low overall performance of GPT-4o mini deems it unsuitable for the bias detection task.

Next, we compare the best BERT model and a baseline from our prior work~\cite{salavati2024reducing} with a fine-tuned TinyLlama. Table~\ref{tab:full-scope results} shows the comparison results. Although TinyLlama achieves high precision, its lower recall causes it to be outperformed by RoBERTa and by the baseline with respect to both F1 and (especially) F2 scores. RoBERTa and the baseline are statistically tied with the highest AUCs ($0.927 \pm 0.003$ and $0.923 \pm 0.004$), indicating that for either model the classification threshold can be tuned to find a trade-off between precision and recall suitable for the target application.

Last, we conduct an ablation test to assess the impact of WSD for data refinement by comparing the performance of RoBERTa and TinyLlama across various dataset configurations. The results in Table~\ref{tab:llm_wsd_results} show that the LN+XN* setting led to higher recall averages than LN+XN (despite not statistically significant) at a small cost in precision. LN achieves the highest recall, but at a steep cost in precision. Therefore, LN+XN* results in the highest F2 scores, indicating that it is the most adequate setting for TAR (Technology Assisted Review) purposes.

\section{Conclusion}
Despite recent strides in fairness, accountability, and transparency, health-related applications and recommender systems are still prone to biases amplified through data, which can perpetuate health disparities and affect patient care. To mitigate this issue, this paper introduces a framework for detecting and diagnosing bias in the medical curriculum, focusing on the data guiding these models rather than on the models' architecture. We use models trained and tested on instructional content annotated by medical experts for bias. We focus on bias related to sex/gender, race/ethnicity, age, and geography. Our method involves extracting non-annotated samples that contain a social identifier as negative samples for the bias classifier. For those extracted negatives, we employ word sense disambiguation to clean out any that have race/ethnicity-related terms but are not actually related to those categories. 

Our findings demonstrate that while LLMs can handle many tasks, they are not well-suited for this one. Our zero- and few-shot prompting with GPT-4o mini underperformed compared to the baseline model from our previous work and scored significantly lower than the language models we tested. Similarly, using a domain-specific model like BioBERT showed no significant improvement. RoBERTa and TinyLlama were the best performers for bias detection, with RoBERTa matching the baseline and showing slight gains in precision and F1 score.

Our WSD models were highly effective at distinguishing biased excerpts from non-biased ones. ALBERT and GlossBERT nearly perfectly disambiguated sentences with race and ethnicity-related keywords. Although GPT models were comparable to BERT models, BERT consistently outperformed GPT in all metrics except recall. While this task focused on one bias category, these models could be adapted to other types with appropriate annotations. Applying WSD to bias detection in medical curricula yielded mixed results. The AUC for RoBERTa was similar to the baseline, but WSD improved both precision and F1 score.

This work could help identify potentially biased excerpts in medical curricula for review before they're used to train models for future health-related applications and recommender systems, contributing to more equitable healthcare across all demographics.

\section{Discussion}
Despite the encouraging results provided by our WSD and bias classification models, there are future directions we can take to enhance our project's significance. First,  in the WSD experiment, using ChatGPT-4o to generate more sentences noticeably increased the performance of our language models. Hence, it is likely that increasing the number of synthetic sentences can further enhance performance if the samples are diverse enough. LLMs often have a ``temperature'' parameter that can control the amount of randomness in the text generation. However, excessively high temperatures could also yield less coherent sentences. 


We also want to consider how word sense disambiguation might be useful in the context of other social identifiers, such as geography (e.g., ``American Heart Association'' vs. ``Native Americans'') and other domains where the tone of an excerpt is more important when evaluating word sense (e.g., social media).

Additionally, although LLMs like GPT models have significantly shown to be advanced in natural language processing, they also present a series of challenges~\cite{naveed2023comprehensive}. Firstly, developing and training LLMs requires computational cost and can be time-consuming. So, they may be less accessible for smaller groups of researchers.



\section{Acknowledgements}
This material is based upon work supported in part by
the National Science Foundation REU Site Grant 2349370 and the WPI STAR Program. 
Any opinions, findings, conclusions, or recommendations expressed in this material are those of the author(s) and do not necessarily reflect the views of the National Science Foundation.

\bibliographystyle{plain} 

\end{document}